\documentclass{bmvc2k}

\title{Pseudo Labelling for Enhanced Masked Autoencoders}

\addauthor{Srinivasa Rao Nandam}{}{1}
\addauthor{Sara Atito}{}{1,2}
\addauthor{Zhenhua Feng}{}{2}
\addauthor{Josef Kittler}{}{2}
\addauthor{Muhammad Awais}{}{1,2}

\usepackage{bbm}
\usepackage{pifont}
\newcommand{\cmark}{\ding{51}}%
\newcommand{\xmark}{\ding{55}}%

\usepackage{multirow}
\usepackage{subcaption}
\usepackage{graphicx}

\addinstitution{
 Surrey Institute for People-Centred AI, University of Surrey, Guildford, GU2 7XH, UK
}
\addinstitution{
 Centre for Vision, Speech and Signal Processing (CVSSP), University of Surrey
}

\runninghead{Student, Prof, Collaborator}{Pseudo Labelling for Enhanced MAE}

\begin{document}

\maketitle

\begin{abstract}
Masked Image Modeling (MIM)-based models, such as SdAE, CAE, GreenMIM, and MixAE, have explored different strategies to enhance the performance of Masked Autoencoders (MAE) by modifying prediction, loss functions, or incorporating additional architectural components. 
In this paper, we propose an enhanced approach that boosts MAE performance by integrating pseudo labelling for both class and data tokens, alongside replacing the traditional pixel-level reconstruction with token-level reconstruction. 
This strategy uses cluster assignments as pseudo labels to promote instance-level discrimination within the network, while token reconstruction requires generation of discrete tokens encapturing local context.
The targets for pseudo labelling and reconstruction needs to be generated by a teacher network. 
To disentangle the generation of target pseudo labels and the reconstruction of the token features, we decouple the teacher into two distinct models, where one serves as a labelling teacher and the other as a reconstruction teacher. This separation proves empirically superior to a single teacher, while having negligible impact on throughput and memory consumption. 
Incorporating pseudo-labelling as an auxiliary task has demonstrated notable improvements in ImageNet-1K and other downstream tasks, including classification, semantic segmentation, and detection. 
\end{abstract}

\section{Introduction}
\label{sec:intro}

Masked Language modeling (MLM), introduced in Bert~\cite{bert}, has demonstarted that masking and predicting the masked tokens provides an effective pretext task for language. Maksed image modelling (MIM) introduced in SiT~\cite{sit} has shown that masking large proportions of image regions/tokens and reconstructing the masked regions provides a strong pretext task that enables self-supervised pretraining (SSP) in vision to outperform supervised pretraining (SP) and works well even in low data regime. BEiT~\cite{beit}, MAE~\cite{mae}, SimMIM~\cite{simmim} and several others \cite{sdae,greenmim,mixmae} continued the idea of MIM, either at the pixel level or the token level, to develop state-of-the-art models that are effective across different downstream tasks.

Masked Auto Encoders~\cite{mae} employ a sparse encoder where the masked regions are dropped and corresponding mask tokens are added only in decoder, typically consisting of 8 transformer blocks' to reconstruct the original image. The sparsity of the encoder makes MAE particularly appealing for pretraining of large and huge vision transformers. Consequently, various studies \cite{sdae, cae, evolvedmae} have investigated various approaches to further enhance its performance. SdAE \cite{sdae} replaces pixel level reconstruction of MAE with token level reconstruction. Evolved MAE \cite{evolvedmae} propose to adopt masking based on attention to further improve representations with MAE. CAE~\cite{cae} adopts an auxiliary task of mask representation prediction which is similar to token level prediction in addition to reconstruction at pixel level utilised by MAE. To the best of our knowledge addition of auxiliary task which introduces a pseudo labelling task to MAE with dropping of tokens in encoders has not been yet explored.
However, the addition of pseudo labelling or instance level discrimination with MIM methods without dropping of masked tokens for encoder has been explored by previous works \cite{sit, mcssl}. 
SiT~\cite{sit} explored the incorporation of infoNCE~\cite{infonce} loss to the MIM pretext task, while MCSSL~\cite{mcssl} investigated the addition of DINO loss on patches in conjunction with MIM through pixel level reconstruction and group masked model learning (GMML)~\cite{gmml}. 
Unlike MAE \cite{mae}, these methods do not employ sparse encoders and utilise large amount of mask tokens (between $50\%$ to $70\%$) for masking, which leads to inefficiencies in the pre-training stage.
In contrast, MAE avoids this inefficiency by dropping the masked tokens, which is a significant portion of the input image ($\sim75\%$).

\begin{figure}[t!]
\begin{center}
\includegraphics[width=\linewidth]{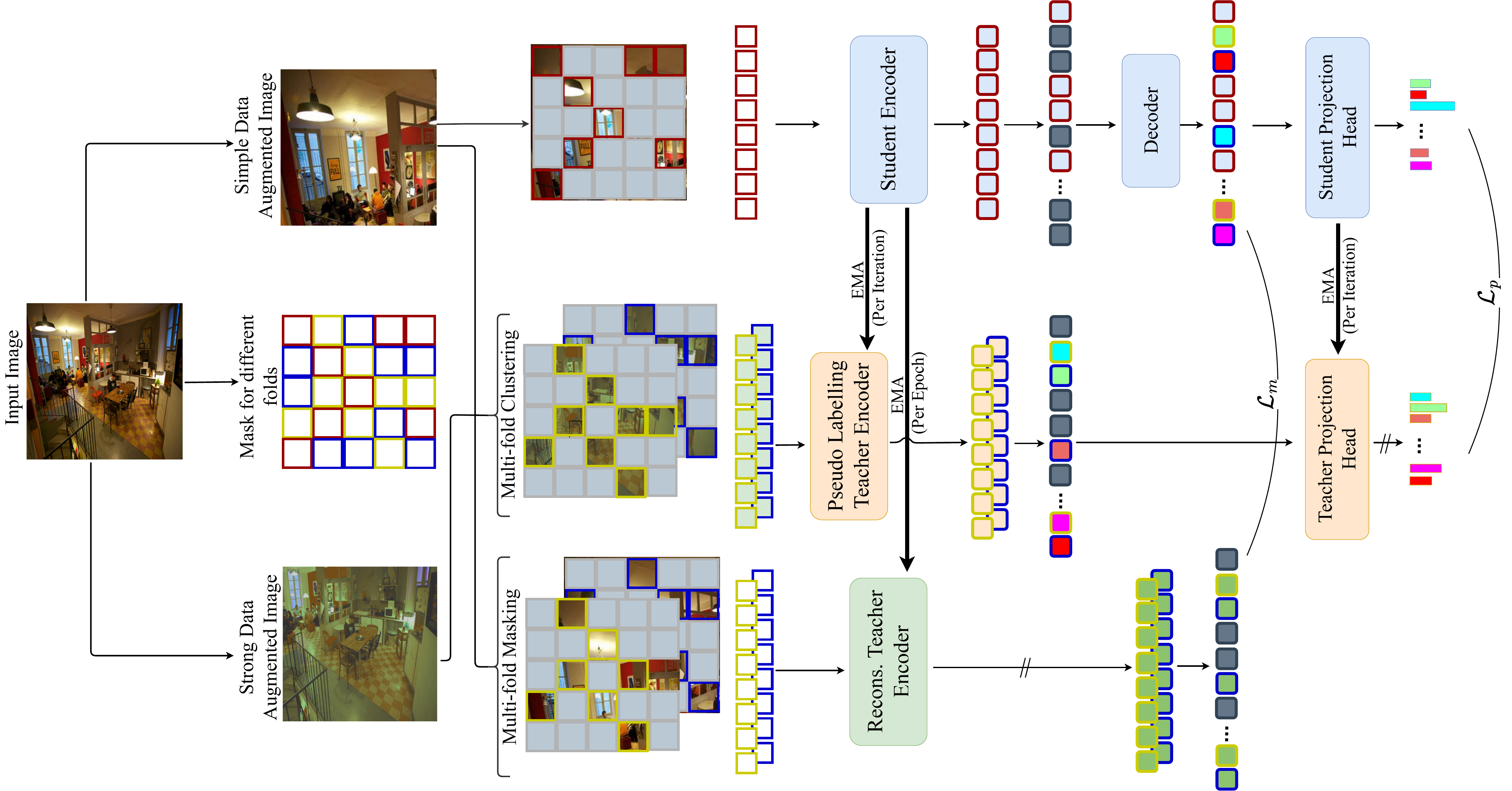}
\end{center}
\caption{The proposed architecture consists of a student encoder, a decoder, and two disentangled teacher encoders updated using EMA. The student encoder processes unmasked regions while the masked tokens are reintroduced at the student decoder. Input image tokens are shuffled via random permutation, with each quarter allocated differently. Multifold pseudo labelling loss is applied to unaligned patch tokens of student and teacher based on feature similarity. The class token pseudo labelling loss is omitted for simplicity.
\vspace{-0.2cm}
}
\label{fig:arch_overview}
\end{figure}

In this work, we introduce a novel approach to Masked Image Modeling that combines the strengths of existing methodologies into a new framework. Our model features a sparse encoder similar to MAE, yet performs reconstruction in the token embedding space, akin to strategies used in \cite{cae, sdae}. This design choice enables more contextually rich information capture. 
We further enhance our model's discriminative capability by integrating an auxiliary pseudo-labeling pretext task. Pseudo labels are generated through clustering by a dedicated pseudo-labeling teacher, enriching the model with nuanced, discriminative information.

Distinctively, our approach diverges from other multi-pretext task models such as \cite{sit, mcssl}, and traditional instance discrimination methods \cite{ibot, dino, swav, msn}, by employing two specialised teachers for each pretext task. The employment of dual teachers effectively disentangles the outputs from each task, ensuring cleaner, more distinct learning signals. Additionally, we utilize different augmented views for different pretext tasks, sourced from two distinct augmentation strategies: one simple~\cite{mae} and one complex~\cite{dino}. This leads to misalignment between the inputs of the pseudo labelling teacher and the student. To resolve this, we address the issue by selecting the most similar patch in the feature space between the teacher and the student. The cluster from the most similar teacher patch is then considered as the target for clustering loss for each masked student cluster prediction.
Inspired from SdAE~\cite{sdae}, which introduces multifold masking to maintain similar mutual information between teacher and student similar, our model extends this strategy to the pseudo-labeling task. This adaptation demonstrated a noticeable improvement in the downstream tasks.
Figure \ref{fig:arch_overview} provides a comprehensive overview of our innovative Pseudo MAE architecture, which effectively and efficiently capturing both fine-grained details and descriminative features.


\section{Related Work}
\label{sec:related_work}
Recent methods for SSL are based on either contrastive/clustering or MIM methods. Contrastive/Clustering methods focus on global semantic representation and can be broadly classified as instance discrimination methods. BERT~\cite{bert} introduced the concept of masking random tokens in a transformer input and predicting those masked tokens for language. For computer vision the first MIM method GMML, which outperformed supervised pretraining, has been introduced in SiT~\cite{sit}. It considers random groups of tokens from an image for masking and predicts the representation in pixel space. Following GMML, SimMIM~\cite{simmim} also uses MIM in pixel space with a light weight decoder. MAE~\cite{mae} explored the concept of sparse encoder where the masked tokens are dropped and reconstruction at pixel level is done with the help of a deep transformer based decoder by reintroducing dropped tokens. 
BeiT~\cite{beit} explored MIM through reconstruction at token level using codebook. The codebook for the target token representation is generated by a pretrained dVAE which has to be trained separately. However, BeiT could not outperform MIM methods which used simple pixel level reconstruction. For large and huge ViTs MAE is compute efficient compared to other methods like \cite{sit, beit} because of sparse processing of only visible tokens in encoder. 

Several methods \cite{sdae, cae, evolvedmae, mixmae, greenmim, ummae} have proposed enhancements to MAE by adding different components or applying it to various hierarchical architectures. UM-MAE~\cite{ummae} proposes a two stage masking strategy for making MAE pretraining viable for hierarchical architectures like PVT\cite{pvt}, Swin\cite{swin}. GreenMIM\cite{greenmim} applies MAE on hierarchical models with the help of a new window attention mechanism and dynamic programming. MixMAE\cite{mixmae} proposes an simple strategy for applying MAE to hierarchical models by considering mixing of multiple images as masking. CAE\cite{cae} separates the pretraining tasks from learning representations with the help of an additional alignment module. \cite{evolvedmae} proposes a masking strategy based on the attention maps of the encoder to capture better semantics. This new masking strategy learns objects parts effectively further enhancing MAE. 

Addition of auxiliary contrastive task to enhance MIM has been studied even in the first successful ViT based MIM model SiT~\cite{sit}. SiT~\cite{sit} proposed GMML which reconstructs the image in original pixel space and considers addition of constrastive loss \cite{simclr} as an auxiliary task to enhance the representations. They aim to capture finegrained information with the GMML and global information with contranstive auxiliary task. MCSSL~\cite{mcssl} propose a framework which captures information at different levels. They use GMML to capture finegrained information and extend DINO~\cite{dino} based clustering loss to patches to capture local semantics. iBoT~\cite{ibot} proposes to extend DINO to patches but does not reconstruct masked images at pixel level. 

Different from these methods Pseudo MAE is different from other works by introducing auxiliary pseudo labelling pretext task to a sparse encoder based MAE where the main pretext task is MIM. We also change reconstruction target from pixels to tokens. We also extend the concept of multifold masking and propose multifold pseduo labelling. We also differ from instance discrimination based methods by different type of augmentations for both teacher and student. Differently from MCSSL and iBoT our method does not need multiple local and global crops and hence the training speed is significantly higher while maintaining performance advantage.

\section{Methodology}
\label{sec:methodology}
Our method is an auto-encoder model where the sparse student encoder is a ViT \cite{vit} and the decoder consists of a few transformer blocks. The student decoder generates token embeddings at the output of the decoder which are used directly for masked reconstruction loss $\mathcal{L}_m$ in feature space. The same token embeddings at the output of the decoder are then passed through projection head to generate pseudo labels. The targets for reconstruction and pseudo labelling are generated by two disentangled teacher encoders. The pseudo labelling teacher generates pseudo labels from the outputs of a projection head, and the token teacher acts as a code-book which generates target tokens. The input to the pseudo labelling teacher and reconstruction teacher are generated from different augmentations of the original input image. The student and teacher share similar input view which is generated from simple data augmentation used by MAE~\cite{mae}. The input view for pseudo labelling teacher is generated from a different crop of the original input based on harsher data augmentations used in DINO~\cite{dino}. We use a multifold masking strategy introduced by SdAE \cite{sdae} to reconstruction teacher, in addition we introduce a multifold class and patch level pseudo labelling which compares class pseudo labels across multiple folds. The class level pseudo labelling enables the network to learn instance level discrimination which is absent in other MIM based methods like MAE\cite{mae}, SimMIM \cite{simmim}. Our patch level pseudo labelling learns local patch level discrimination which is different from our token level reconstruction which captures global shape information similar to SdAE \cite{sdae}. We discuss the framework and the different components in detail in the following sections.

\subsection{Masked Image Modelling}

Masked Image Modelling (MIM) has emerged as a significant self-supervised learning approach in recent research \cite{sit, mae, simmim, gmml, sdae}, showcasing its effectiveness and broad utility in the domain of image processing. At its core, MIM revolves around the reconstruction of an original image from an input image that has been partially masked. In this section, we provide an in-depth discussion of our novel MIM approach, utilising vision transformers.

Consider an input image $\mathbf{X}_{in} \in \mathcal{R}^{H \times W \times C}$, where $H$, $W$, and $C$ denote height, width, and the number of channels, respectively. We generated two different $\mathbf{X}_{simp}, \mathbf{X}_{comp}$ views based on data augmentations used in MAE~\cite{mae} and DINO~\cite{dino} respectively. Both reconstruction teacher and student use the $\mathbf{X}_{simp}$ as their input, whereas pseudo labelling teacher uses $\mathbf{X}_{comp}$. The image is then converted into a sequence of flattened patches, resulting in a two-dimensional matrix $\mathbf{X}_{simp} \in \mathcal{R}^{N \times (P^{2}.C)}$, where $N$ is the number of patches and $P\times P$ is the resolution of each patch. 
Subsequently, we generate a random mask $\textbf{M} ={(m_{1}, \dots, m_{N}})$ for the $N$ tokens, with each $m_{i}\in{\{0, 1\}}$, determining the tokens to be retained or discarded. 

For the student, we generate the masked input $\mathbf{\hat{X}} \in \mathcal{R}^{ (N'+1) \times (P^{2}.C)}$, where $N'$ is the number of the tokens corresponding to $m_{i}=0$ including the class token. The output of the student after passing through the encoder and decoder is $\mathbf{\tilde{X}\in R}^{(N'+1) \times D}$, where $D$ is the output embedding dimension. The objective function for reconstruction is simply a cosine similarity loss between targets and decoder output embeddings.

\begin{equation} \label{eq:recons}
    \mathcal{L}_{m} = \frac{1}{N'}\sum_{i}^{N} cos\_sim( m_{i} \times (\overline{t_{rec}(\mathbf{X}^u_i)}, \mathbf{\tilde{X}}_i )
\end{equation}

\begin{equation} \label{eq:t_norm}
    \overline{t_{rec}(\mathbf{X}_i)} = \frac{t_{rec}(\mathbf{X}_i)-mean(t_{rec}(\mathbf{X}_i))}{\sqrt{var(t_{rec}(\mathbf{X}_i))+\epsilon}}
\end{equation}

Equation \ref{eq:recons} gives the equation for reconstruction loss where $t_{rec}$ is the reconstruction teacher encoder generating target tokens. Here $\mathbf{X}^{u}$ is the input to the reconstruction teacher. This is different from MAE which uses original input pixels as the targets. Here $\tilde{\mathbf{X}}_{i} = (\tilde{\mathbf{X}}_{1}, \dots, \tilde{\mathbf{X}}_{N})$ is the output corresponding to patches excluding the class token output. We use normalised teacher inputs as the targets which are given by the Equation(\ref{eq:t_norm}).

In addition we use a multifold masking strategy \cite{sdae} for generating targets with teacher encoder. Let $\mathbf{X}_{u}$ represent the input where $\mathbf{X}_{u}=\{\mathbf{X}_{simp}(i) \mid \mathbf{m}_i=0\}$. The input $\mathbf{X}_{u} \in \mathbf{R}^{(N-N') \times (P^{2}.C)}$ is split in $K$ folds ($\mathbf{X}_{u}^{1}, \dots, \mathbf{X}_{u}^{k}$) to which class tokens are prepended to generate multifold split outputs ($(t_{rec}(\mathbf{X}_{u}^{1}), \dots, t_{rec}(\mathbf{X}_{u}^{k})$). The output patches excluding that of class token of these multifold outputs is utilised in Equation(\ref{eq:recons}). 

\subsection{Disentangled Teachers}
Several works like \cite{sit, mcssl, ibot} explore combining their primary pretext task clustering/contrastive learning with auxiliary pretext task of MIM. Apart from their encoders not being sparse due to the usage of mask tokens/zeros, they utilise a single teacher for both clustering/contrastive learning and their auxiliary MIM task which is updated with exponential moving average (EMA). We propose to use two teachers, namely pseudo labelling teacher $t_{cl}$ and reconstruction teacher $t_{rec}$, that disentangle the task of generating target cluster assignments from generating target token representations. The teacher for token representation benefits from having stable teacher where the teacher is updated by EMA less frequently, i.e. per epoch. The teacher for target cluster assignments performs better when updated frequently, i.e. per iteration. Utilising two different teachers specialised in different pretext tasks enables us to also apply two different EMA update schedules with different frequencies. Specifically, the EMA for our pseudo labelling teacher is done for each iteration with a starting momentum of $0.996$ where as the EMA of the reconstruction teacher is done for each epoch with a starting momentum of $0.96$.

\subsection{Multifold Pseudo Labelling}
Multifold masking produces noticeable improvements for downstream tasks as shown in \cite{sdae}. We aim to extend multifold masking strategy for pseudo labelling as well. We generate pseudo labelling unmasked input $\mathbf{X}_{cu}=\{\mathbf{X}_{comp}(i) \mid \mathbf{m}_i=0\}$ from complex data augmented view $X_{comp}$ with mask $m$. Given the pseudo labelling unmasked input $\mathbf{X}_{cu} \in \mathbf{R}^{(N-N') \times (P^{2}.C)}$, we split in $K$ folds ($\mathbf{X}_{cu}^{1}, \dots, \mathbf{X}_{cu}^{k}$) and prepend class token to each fold. The resulting folds are passed through pseudo labelling teacher $t_{cl}$ to generate multifold pseudo labelling split outputs ($(\mathbf{\bar{X}}^{1}, \dots, \mathbf{{\bar{X}}}^{k}$) which include class and patch teacher encoder outputs for each fold. These encoder outputs are then passed through the projection head with a few linear layers and has a final output projection dimension of $4096$ for both patch and class token layers. The pseudo labels are generated as cluster assignments from projection head outputs. We apply sinkhorn normalization to avoid collpase and generate target cluster predictions. 
Let $(\mathbf{P}^{1}, \dots, \mathbf{P}^{k})$ be the target patch pseudo labels and $(\mathbf{C}^{1}, \dots, \mathbf{C}^{k})$ be the target class pseudo labels after projection head for different folds. 

We generate student decoder predictions $\tilde{\mathbf{X}}^{m}=\{\tilde{\mathbf{X}}_i \mid \mathbf{m}_i=0\}$ corresponding to masked regions. $\tilde{\mathbf{X}}^{m}$ is passed through the student projection head to generate  pseudo labels $\mathbf{\tilde{P}}, \mathbf{\tilde{C}}$ corresponding to patch and class tokens respectively. We generate inputs for pseudo labelling and student based on two different augmentations which causes misalignment between these views. This makes applying clustering loss non trivial unlike other clustering methods~\cite{ibot, dino} where similar views are passed through both teacher and student. 
To apply clustering loss on unaligned patches, we find the closest target pseudo label for each student pseudo label predictions. We perform the closest match by finding nearest teacher patch for each student patch in the feature space. Let $\mathbf{\hat{P}}_n=\mathbf{P}^{k}_{i} (\textit{ where } \mathbf{P}^{k}_{i} = min\{dist(\tilde{\mathbf{X}}_n, \mathbf{\bar{X}}_{i}^{k})\})$ be the closest target pseudo label for $n^{th}$ student label prediction. For the final class token target $\mathbf{\hat{C}}=Avg(\mathbf{C}^{k})$ pseudo label, we take the average for all folds.


\begin{equation} \label{eq:ptch}
        \mathcal{L}_{p} = \frac{1}{N-N'}\sum_{n=1}^{N-N'}\text{H}(\mathbf{\hat{P}}_{n}, \mathbf{\tilde{P}}_{n})
\end{equation}

\begin{equation} \label{eq:cls}
        \mathcal{L}_{c} = \text{H}(\mathbf{\hat{C}}, \mathbf{\tilde{C}})
\end{equation}

To calculate the patch level pseudo labelling loss, we apply cross entropy loss $H$ between different teacher patch folds $\mathbf{p}^{i}$ and student patch folds $\mathbf{\tilde{p}}^{i}$ which is given by Equation \ref{eq:ptch}. For class level pseudo labelling loss, we calculate loss between the single student class pseudo label $\mathbf{\tilde{c}}$ with teacher class pseudo label of each fold $\mathbf{c}^{i}$. The Equation \ref{eq:cls} provides formulation for class teacher class pseudo labelling loss. 

\begin{equation} \label{eq:loss}
        \mathcal{L} = \lambda_{m}*\mathcal{L}_{m} + \lambda_{c}*\mathcal{L}_{c} + \lambda_{p}*\mathcal{L}_{p}
\end{equation}

The final loss is given by Equation \ref{eq:loss}, is the combination of reconstruction loss $\mathcal{L}_{m}$, class level pseudo labelling loss $\mathcal{L}_{c}$, and patch level pseudo labelling loss $\mathcal{L}_{p}$ with scaling factors $\lambda_{m}, \lambda_{c}, \lambda_{p}$ respectively set to $1$ by default.

\begin{table}[htbp]
\begin{center}
\begin{tabular}{ccc}
\subfigure{\includegraphics[width=0.3\textwidth]{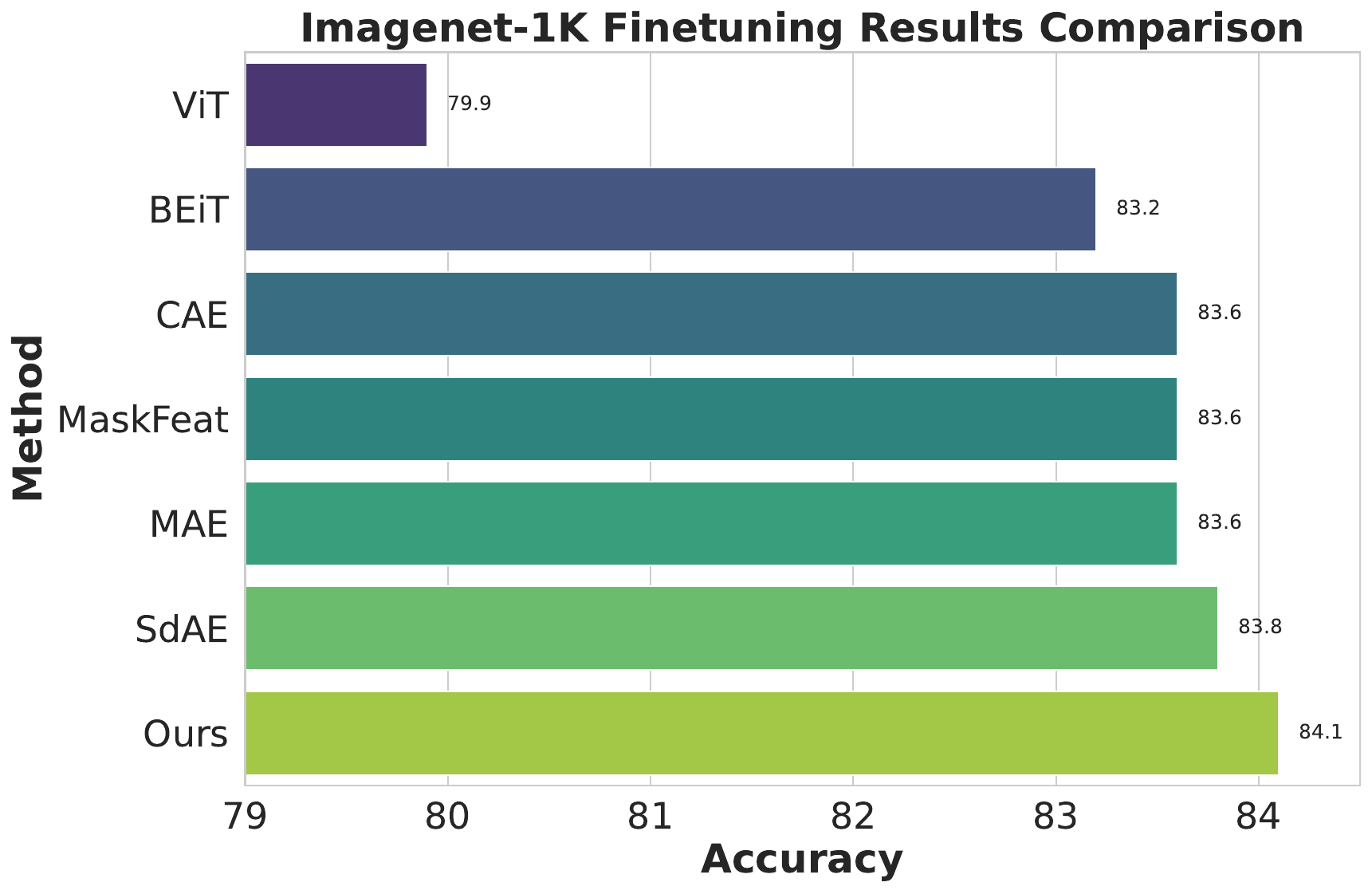}}&
\subfigure{\includegraphics[width=0.3\textwidth]{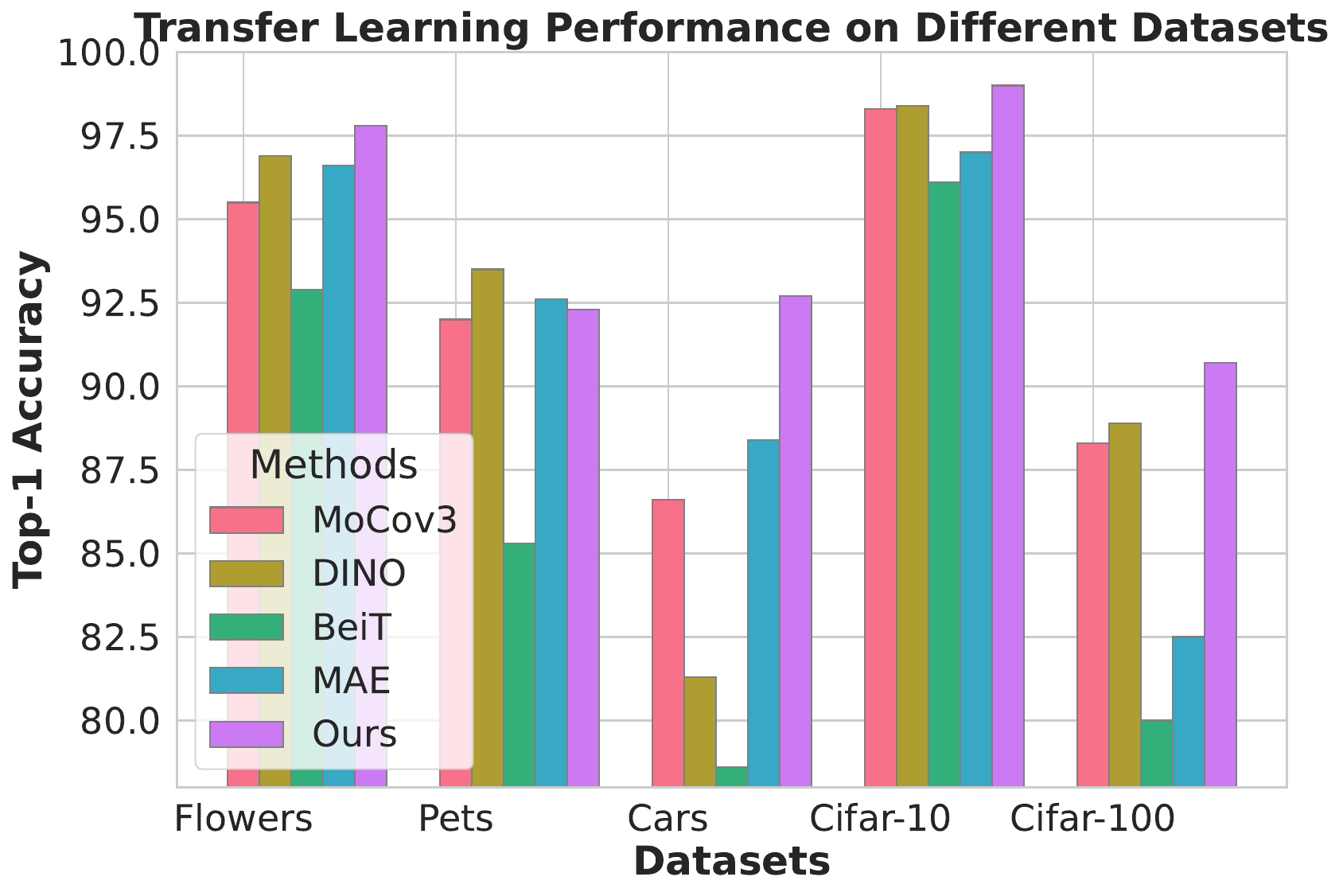}} &
\subfigure{\includegraphics[width=0.3\textwidth]{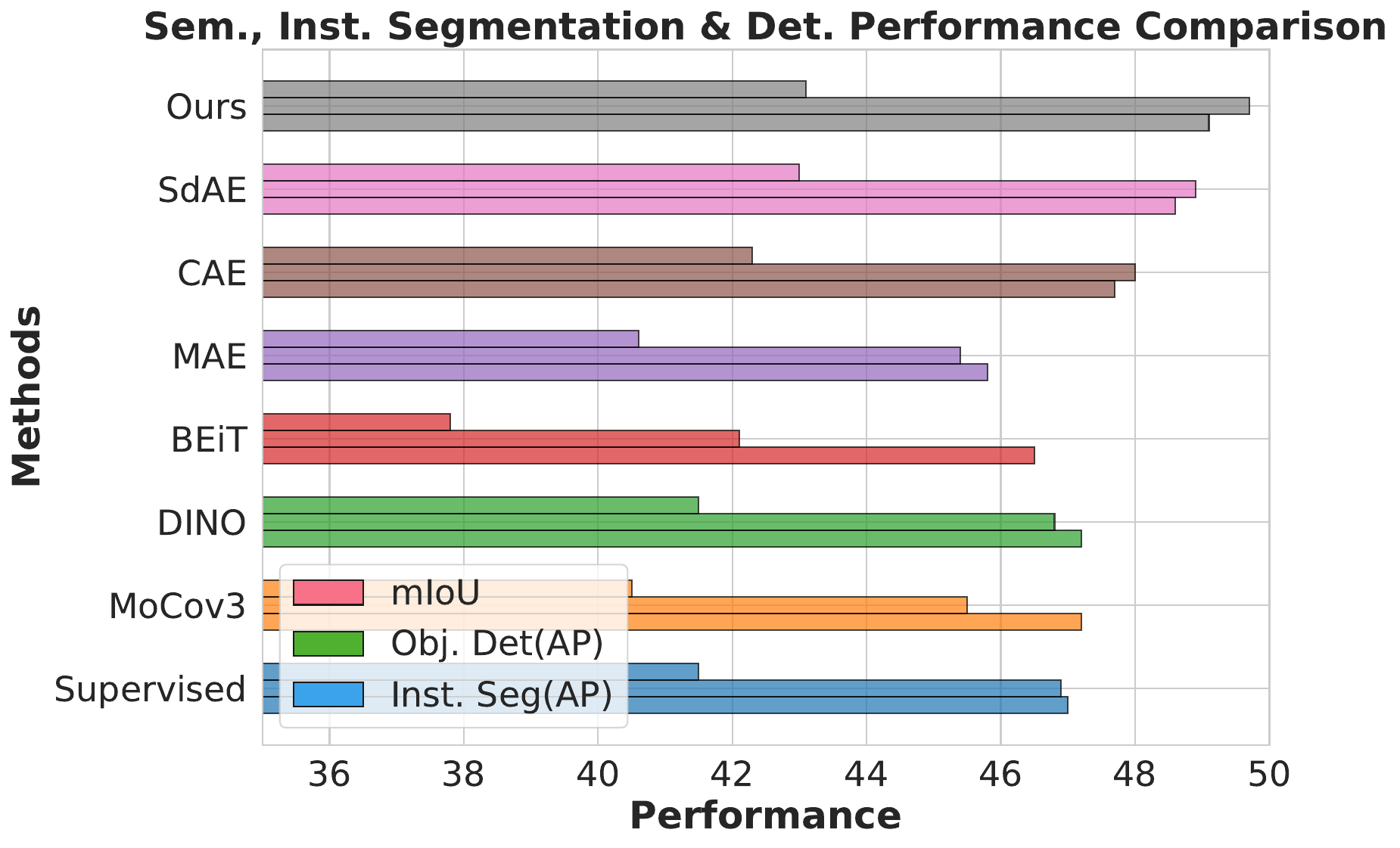}} \\
\end{tabular}
\end{center}
\caption{Plots for (a)ImageNet classification, (b)Transfer Learning, (c)Det. \& Seg.}
\label{table:plots}
\end{table}

\begin{table}
\parbox{.49\linewidth}{
\begin{center}
\resizebox{0.99\linewidth}{!} {
\begin{tabular}{l | c c c c c c }
 \hline
 Method & Backbone & Supervision & Pretrain Epochs & Finetuning\\
 \hline
 ViT\cite{vit} & ViT-B & RGB & - & 79.9\\  
 BEiT\cite{beit} & ViT-B & DALL-E & 800 & 83.2\\
 CAE\cite{cae} & ViT-B & DALL-E & 800 & 83.6\\
 MaskFeat\cite{maskfeat} & ViT-B & HOG & 300 & 83.6\\
 iBOT\cite{ibot} & ViT-B & Momentum & 1600 & 84.0\\
 MCSSL\cite{mcssl} & ViT-B & Momentum & 800 & 84.0\\
 MAE\cite{mae} & ViT-B & Momentum & 800 & 83.6\\
 SdAE*\cite{sdae} & ViT-B & Momentum & 300 & 83.8\\
 Ours & ViT-B & Momentum & 300 & \textbf{84.1}\\
\hline
\end{tabular}
}
\end{center}
\caption{Imagenet-1K finetuning results on various methods compared with ours. * represents our replicated model of SdAE with official pretraining and finetuning code.}
\label{table:imnet}
}
\hfill
\parbox{.49\linewidth}{
\begin{center}
\resizebox{0.99\linewidth}{!} {
\begin{tabular}{l  c c c c c c} 
 \hline
 Method & Backbone & Flowers & Pets & Cars & Cifar-10 & Cifar-100 \\ 
 \hline 
MoCov3 & ViT-B & 95.5 & 92.0 & 86.6 & 98.3 & 88.3 \\ 
DINO & ViT-B & 96.9 & 93.5 & 81.3 & 98.4 & 88.9 \\
BeiT & ViT-B & 92.9 & 85.3 & 78.6 & 96.1 & 80.0 \\
MAE & ViT-B & 96.6 & 92.6 & 88.4 & 97.0 & 82.5 \\
Ours & ViT-B & \textbf{97.8} & 92.3 & \textbf{92.7} & \textbf{99.0} & \textbf{90.7} \\ 
 \hline
\end{tabular}
}
\end{center}
\caption{Transfer learning by finetuning pretrained models with the ViT-B/16 backbone on diverse datasets. We report top-1 accuracy and the results for different methods are directly from~\cite{mixaf}.}
\label{table:small_finetune}
}
\end{table}

\section{Experiments}
\label{sec:experiments}
We base our experiment setup on pretraining first and finetuning next to evaluate our approach across several downstream tasks and datasets. We pretrain our model based on Imagenet-1K\cite{imagenet} following previous works\cite{dino, ibot, mae}. We evaluate our model on multiple downstream tasks like multi class classification, semantic segmentation, objection detection, instance segmentation. The pretraining for Imagenet-1K is done for 800 epochs with AdamW\cite{adamw} as the optimiser. The default image size is set to $224\times 224$ and the patch size for ViT\cite{vit} is set to $16$. The masking ratio is set to $75\%$ following MAE, the learning rate is set to $2.666e-4$ and warmup epochs to $60$. The staring, ending momentum for momentum schedule of reconstruction teacher is set to $0.96, 0.99$ respectively where the momentum update is done for each epoch. The staring, ending momentum for momentum schedule of pseudo labelling teacher is set to $0.996, 1$ respectively where the momentum update is done for each iteration. More ablations and visualisation to be provided in the supplementary section. Performance plots provided in Table~\ref{table:plots}

\subsection{Results on multi class classification}
We evaluate our model on Imagenet-1K\cite{imagenet} by finetuning for $100$ epochs on ViT-B in Table~\ref{table:imnet}. We set a base learning rate of $5e-4$  warmup epochs to $5$.  We utilise an effective batch size of $1024$ with weight decay set to $0.05$ and layer decay to $0.65$. We compare our model to other SSL methods like DINO\cite{dino}, BEiT\cite{beit}, MAE\cite{mae} on finetuning. Interestingly we show better performance than MAE~\cite{mae} with only $300$ epochs compared to MAE with $1600$ epochs. We also outperform iBoT~\cite{ibot} with out requiring large number of local crops~(10) and less epochs of training. We also evaluate on other smaller datasets like Pets, Cars, Cifar10, Cifar100. The finetuning is done for $1000$ epochs with a similar settings to DINO\cite{dino} in Table~\ref{table:small_finetune}. When compared against other methods we find that our method performs better thus excelling at both large and smaller datasets.

\subsection{Results on semantic segmentation, object detection, instance segmentation}
We perform the finetuning and evaluation for semantic segmentation on ADE20K dataset with results provided in Table~\ref{table:sem_seg}. We follow previous SSL approaches\cite{dino, moco, simclr} by utlising Upernet\cite{upernet} as the task layer. We follow MAE\cite{mae} for setting the schedule and learning rate. Semantic segmentation shows the capability of the network to capture semantic details at various levels and provides for a good downstream task. Our models excels other competing models with less number of epochs for the complex task of semantic segmentation. Following MAE\cite{mae} we finetune our architecture on COCO dataset\cite{mscoco} with results in Table~\ref{table:obj_det}. We use Mask R-CNN\cite{maskrcnn} and we also apply FPN\cite{fpn} to train the network. The learning rate and the schedule used follows MAE\cite{mae}. Our method excels the competing methods in object detection and instance segmentation showcasing its capability to capture mutiple instances.

\begin{table}
\parbox{.49\linewidth}{
\begin{center}
\resizebox{0.99\linewidth}{!} {
\begin{tabular}{l  c c c c c} 
 \hline
 Method & Backbone & Pretrain Epochs & FLOPs (G) & Params (M) & mIoU \\ 
 \hline 
 Supervised & ViT-B & 300 & 606 & 164 & 47.0\\
MoCov3 & ViT-B & 300 & 606 & 164 & 47.2\\
DINO & ViT-B & 400 & 606 & 164 & 47.2\\
BEiT & ViT-B & 300 & 606 & 164 & 45.5\\
BEiT & ViT-B & 800 & 606 & 164 & 46.5\\
MAE & ViT-B & 300 & 606 & 164 & 45.8\\
MAE & ViT-B & 1600 & 606 & 164 & 48.1\\
CAE & ViT-B & 300 & 606 & 164 & 47.7\\
SdAE & ViT-B & 300 & 606 & 164 & 48.6\\
Ours & ViT-B  & 300  & 606 & 164 & \textbf{49.1}  \\ 
 \hline
\end{tabular}
}
\end{center}
\caption{Downstream task evaluation on ADE20K semantic segmentation. We compare our method against other SSL methods and the evaluation settings follow a standard approach.}
\label{table:sem_seg}
}
\hfill
\parbox{.49\linewidth}{
\begin{center}
\resizebox{0.99\linewidth}{!} {
\begin{tabular}{l  c c c c c c c c} 
 \hline
 \multirow{2}{*}{Method} &  \multirow{2}{*}{Backbone} & \multirow{2}{*}{Pretrain Epochs} & \multicolumn{3}{c|}{Object detection} & \multicolumn{3}{c}{Instance Segmentation}\\ 
  &  &  & $AP^{b}$ & $AP^{b}_{50}$ & \multicolumn{1}{c|}{$AP^{b}_{75}$}  & $AP^{m}$ & $AP^{m}_{50}$ & $AP^{m}_{75}$\\ 
 \hline 
 Supervised & ViT-B & 300 & 46.9 & 68.9 & 51.0 & 41.5 & 65.5 & 44.4\\
MoCov3 & ViT-B & 300 & 45.5 & 67.1 & 49.4 & 40.5 & 63.7 & 43.4\\
DINO & ViT-B & 400 & 46.8 & 68.6 & 50.9 & 41.5 & 65.3 & 44.5\\
BEiT & ViT-B & 300 & 39.5 & 60.6 & 43.0 & 35.9 & 57.7 & 38.5\\
BEiT & ViT-B & 800 & 42.1 & 63.3 & 46.0 & 37.8 & 60.1 & 40.6\\
MAE & ViT-B & 300 & 45.4 & 66.4 & 49.6 & 40.6 & 63.4 & 43.7\\
MAE & ViT-B & 1600 & 48.4 & 69.4 & 53.1 & 42.6 & 66.1 & 45.9\\
CAE & ViT-B & 300 & 48.0 & 68.7 & 52.7 & 42.3 & 65.6 & 45.4\\
SdAE & ViT-B & 300 & 48.9 & 69.6 & 53.3 & 43.0 & 66.2 & 46.2\\
Ours & ViT-B  & 300 & \textbf{49.7} & 68.1 & 54.0 & \textbf{43.1} & 65.4 & 46.7\\ 
 \hline
\end{tabular}
}
\end{center}
\caption{Downstream task evaluation on COCO object detection and instance segmentation. We compare our method against other SSL methods and the evaluation settings follow a standard approach.}
\label{table:obj_det}
}
\end{table}

\section{Ablation}
\label{sec:ablation}
We conduct ablation study on Imagenet-1K with ViT as the backbone. We study the usage of multiple teachers for different pretext task. We explore the effect of EMA having different update frequencies i.e. per epoch or per iteration on each of the teacher. We also explore the effect of longer pretraining and different losses in our architecture. Augmentations are generally used in all clustering frameworks\cite{dino, mcssl, sit}. We explore the effect of simple and complex data augmentations in the context of pseudo labelling pretext task. We use ViT-S for all the experiments except for Table~\ref{table:ml_clust}. The pretraining and finetuning is done of Imagenet-1K for $100$ and $50$ epochs respectively. For Table~\ref{table:ml_clust} we use ViT-B where we pretrain, finetune for $300$, $100$ epochs respectively.

\begin{table}
\parbox{.49\linewidth}{
\begin{center}
\resizebox{\linewidth}{!} {
\begin{tabular}{c | c c c}
 \hline
 Method & In-1K\% & Pretraining Time & Memory\\
 \hline
 SdAE~\cite{sdae} & 74.19 & 2.5 days & 11.2G\\
 MAE~\cite{mae} & - & 12.7 days & 10G\\ 
Single Teacher(Per iteration) & 73.5 & 2.9 days & 12G \\  
Single Teacher(Per epoch) & 74.3 & 2.9 days &  12G \\
Disentangled Teacher(default) & 74.8 & 3.2 days & 14.5G\\
\hline
\end{tabular}
}
\end{center}
\caption{Results of our method compared to a single teacher which generates both pseudo labels and target tokens~(with ViT-S).  We also provide pretraining time and memory consumption for ViT-B.}
\label{table:ab_diff_teachers}
}
\hfill
\parbox{.49\linewidth}{
\begin{center}
\resizebox{0.99\linewidth}{!} {
\begin{tabular}{c | c c c}
 \hline
 Backbone & Initial Momentum & Final Momentum & In-1K\\
 \hline
 ViT-S & 0.96 & 1 &  73.6\\  
 ViT-S & 0.996 & 1 & 74.8\\
  \hline
\end{tabular}
}
\end{center}
\caption{Results of using different momentum's for our cluster teacher. Large momentum's work better for generating pseudo labels.}
\label{table:ab_momentum}
}
\end{table}

\noindent
\textbf{Effect of Multiple Teachers and EMA update frequency, schedule}
We compare a single teacher to disentangled ones (Table~\ref{table:ab_diff_teachers}). Teachers update per epoch or iteration. Our disentangled method, with one teacher for pseudo-labeling and another for reconstruction, outperforms single teachers. Using two teachers allows separate EMA schedules for stable training. We analyze memory and pretraining time for different teachers on ViT-B with $300$ epochs and batch size $768$, finding dual teachers marginally increase time and memory. Unlike MAE~\cite{mae}, our approach reduces pretraining time with slightly more memory.

We explore momentum schedules for the Pseudo labeling teacher in Table~\ref{table:ab_momentum}. As SdAE~\cite{sdae} studied momentum for the reconstruction teacher, we focus on schedules with pseudo labelling teacher with initial momenta of $0.96$ and $0.996$, reaching $1$. The $0.996$ schedule performs better, showing the need for frequent, smaller updates. This highlights the importance of disentangled teachers for different pretext tasks.

\begin{table}
\parbox{.45\linewidth}{
\begin{center}
\resizebox{0.8\linewidth}{!} {
\small
\begin{tabular}{c | c c c c}
Backbone & $\lambda_m$ & $\lambda_c$ & $\lambda_p$ & In-1K\\
 \hline
 ViT-S & 1 & 0 & 0 & 74.19\\  
 ViT-S & 1 & 0 & 1 & 73.20\\
 ViT-S & 1 & 1 & 0 & 74.23 \\
 ViT-S & 0 & 1 & 1 & 74.09\\
 ViT-S & 1 & 1 & 1 & 74.82\\
\end{tabular}
}
\end{center}
\caption{Results of our model for different values of scaling factors $\lambda_m$, $\lambda_c$, $\lambda_p$. All the methods are pretrained for $100$ epochs and finetuned for $50$ epochs. }
\label{table:ab_loss}
}
\hfill
\parbox{.45\linewidth}{
\begin{center}
\resizebox{0.99\linewidth}{!} {
\small
\begin{tabular}{c c c}
 Data Aug. & Multifold Pseudo Lab. & In-1K\\
 \hline
  Simple & \xmark & 83.6\\ 
  Simple & \cmark & 83.8\\  
  Complex & \xmark & 83.9\\
  Complex & \cmark & 84.1\\
\end{tabular}
}
\end{center}
\caption{Table showcasing effect of multifold clustering and data augmentation for pseduo labelling pretext task with ViT-B pretrained for $300$ epochs.}
\label{table:ml_clust}
}
\end{table}

\noindent
\textbf{Effect of different pseudo label loss terms}
We study the effects of different auxiliary pretext pseudo labelling loss terms on the performance of the model in Table~\ref{table:ab_loss}. We find that when we apply only pseudo label loss on patches with i.e. $\lambda_{p}=0$, we get a lower accuracy. We find that when we apply both class and patch level pseudo labelling loss it leads to a slight improvement in performance. This proves that auxiliary task of pseudo labelling is required for better performance.

\noindent
\textbf{Effect of Multifold clustering}
We also study the effects of applying multifold pseudo labelling and the choice of data augmentation in Table~\ref{table:ml_clust}. Multifold masking\cite{sdae} shows that keeping the sufficient mutual information provides for a better encoder. We introduce similar concept to pseudo labelling and study its effects. We find that following multifold pseudo labelling generates better results than with application of normal clustering. In addition we also find that simple augmentation~\cite{mae} does not work with pseudo labelling and it requires complex augmentations~\cite{dino}.

\vspace{-1em}
\section{Conclusion}
\label{sec:conclusion}
We introduce an MIM model with an auxiliary pretext task of pseudo label generation to enhance instance discrimination globally and locally. Initially, adding this task alone doesn't yield extra gains due to the distinct nature of pretext tasks. To address this, we propose two disentangled teachers: one for pseudo labeling and another for token-based reconstruction. Token reconstruction requires a stable teacher, updated with Exponential Moving Average (EMA) at each epoch, while pseudo labeling requires more frequent EMA updates at each iteration. We stress the importance of capturing information at multiple levels for better encoder performance. Our future work aims to extend this to multimodality and study teacher behavior for different multimodal tasks.

\noindent
\textbf{Acknowledgements:}
This work was supported in part by the EPSRC grants MVSE (EP/V0 02856/1) and JADE2 (EP/T022205/1).
\appendix

\section{Implementation Details}
\textbf{Pretraining.} We adopt a pretraining configuration similar to that used by SdAE~\cite{sdae}. Specifically, we utilize the AdamW optimizer and train our model for 300 epochs with a batch size of 768. The learning rate is set to 8e-4 and follows a cosine decay schedule, starting with a 60 epoch warmup. We apply a weight decay of 0.05 and a drop path rate of 0.25 exclusively on the encoder. The momentum coefficient for reconstruction teacher starts at 0.96 and increases to 0.99 according to a cosine schedule, with EMA updated at the end of each pretraining epoch. For pseudo-labeling, the teacher's initial momentum starts at 0.996 and ends at 1, with EMA updates performed at each iteration. We use MAE~\cite{mae} augmentations for both the reconstruction teacher and student, and DINO~\cite{dino} augmentations for the pseudo-labeling teacher. The masking ratio is set to 0.75, meaning only 49 tokens are passed to the student branch. Masked tokens are divided into three groups of 49 tokens each, and these groups are processed by a shared weighted teacher branch.

\textbf{Fine-tuning on ImageNet.} For fine-tuning, we adhere closely to the procedures described in SdAE~\cite{sdae}, employing layer-wise learning rate decay, weight decay, and the AdamW optimizer. The batch size is set to 2048 with a weight decay of 0.05. For the ViT-B model, we conduct 100 training epochs with a base learning rate of 5e-4, a layer-wise decay rate of 0.65, a drop path rate of 0.1, and a 10-epoch warmup. For the ViT-L model, we perform 50 training epochs with a base learning rate of 1e-3, a layer-wise decay rate of 0.75, a drop path rate of 0.2, and a 5-epoch warmup.

\textbf{Transfer Learning on Smaller Datasets.} When transferring to smaller datasets, we follow the configuration details provided for DeiT~\cite{deit}. The training is conducted with a batch size of 512, a learning rate of 5e-4, and weight decay of 0.05. We employ a warmup period of 5 epochs followed by a cosine decay schedule for the learning rate. The training process runs for 100 epochs, with data augmentation techniques such as random cropping, horizontal flipping, and color jittering applied to improve generalization.

\textbf{Object Detection and Instance Segmentation on COCO.} We implement the same settings as iBoT~\cite{ibot} for object detection and instance segmentation. This includes multi-scale training where the short side of the image is resized between 480 and 800 pixels, and the long side is no more than 1333 pixels. We use a batch size of 32, a learning rate of 3e-4, and a layer-wise decay rate of 0.75. Training follows the 1× schedule, but we use 25 epochs with the learning rate reduced by a factor of 10 at epochs 9 and 11. Multi-scale testing is not employed. The Cascade Mask R-CNN model is implemented using MMDetection.

\textbf{Semantic Segmentation on ADE20K.} For semantic segmentation tasks on the ADE20K dataset, we follow the settings detailed in SdAE~\cite{sdae}. The AdamW optimizer is used with a batch size of 16 and a layer-wise decay rate of 0.65. Input images are resized to 512 × 512 pixels. We set the learning rate to 4e-4 for all experiments and fine-tune the model for 160,000 steps, without employing multi-scale testing.

\pagebreak
\bibliography{egbib}
\end{document}